\documentclass{article}

\usepackage{arxiv}

\usepackage[utf8]{inputenc} 
\usepackage[T1]{fontenc}    
\usepackage{hyperref}       
\usepackage{url}            
\usepackage{booktabs}       
\usepackage{amsfonts}       
\usepackage{nicefrac}       
\usepackage{microtype}      

\usepackage{cite}
\usepackage{amsmath,amssymb,amsfonts}
\usepackage{algorithmic}
\usepackage{textcomp}
\usepackage{xcolor}
\usepackage{enumitem}
\usepackage{lipsum,graphicx,subcaption}

\usepackage{wrapfig}
\usepackage{multirow}

\title{Dynamic Neural Diversification: Path to Computationally Sustainable Neural Networks}

\author{

  Alexander Kovalenko \\
  Faculty of Information Technology\\
  Czech Technical University in Prague\\
  Prague, Czech Republic\\
  \texttt{kovalale@fit.cvut.cz} \\
  \And
  Pavel Kordík \\
  Faculty of Information Technology\\
  Czech Technical University in Prague\\
  Prague, Czech Republic\\
  \texttt{pavel.kordik@fit.cvut.cz} \\
  \And
  Magda Friedjungová \\
  Faculty of Information Technology\\
  Czech Technical University in Prague\\
  Prague, Czech Republic\\
  \texttt{magda.friedjungova@fit.cvut.cz} \\
}

\date{}

\renewcommand{\undertitle}{Paper accepted to ICANN 2021.}

\begin{document}
\maketitle

\begin{abstract}
Small neural networks with a constrained number of trainable parameters, can be suitable resource-efficient candidates for many simple tasks, where now excessively large models are used. However, such models face several problems during the learning process, mainly due to the redundancy of the individual neurons, which results in sub-optimal accuracy or the need for additional training steps. Here, we explore the diversity of the neurons within the hidden layer during the learning process, and analyze how the diversity of the neurons affects predictions of the model. As following, we introduce several techniques to dynamically reinforce diversity between neurons during the training. These decorrelation techniques improve learning at early stages and occasionally help to overcome local minima faster. Additionally, we describe novel weight initialization method to obtain decorrelated, yet stochastic weight initialization for a fast and efficient neural network training. Decorrelated weight initialization in our case shows about 40\% relative increase in test accuracy during the first 5 epochs.
\end{abstract}

\keywords{Diversification  \and Negative Correlation \and Weight Initialization \and Computational Sustainability \and Neural Networks}

\section{Introduction}
Over the last decade, machine learning algorithms have achieved vast progress in various fields. Namely, general approach called deep neural networks (DNN) with multiple hidden layers \cite{hinton2006reducing}, has enabled machine learning algorithms to perform at an acceptable level in the many areas, in some cases outperforming human accuracy \cite{ciresan2011flexible}. Such progress, in no small measure, has become available due to modern hardware computational capabilities, enabling the training of large DNN on an immense amount of data.

On the other hand, even though large models perform very well on complex tasks, we cannot endlessly rely on an infinite increase in computational resources and size of datasets. Training large neural networks is energy, time and memory demanding task. Recently, researchers started questioning energy consumption of machine learning algorithms and their carbon footprint \cite{lacoste2019quantifying}. Thus it will not be superfluous to develop a strategy for the models that have a constrained number of parameters, sufficient enough for the certain task, and can be trained fast, rather than chasing higher accuracy by enlarging the number of parameters and using more complex hardware.

Universal approximation theorem \cite{cybenko1989approximation} claims that a feed-forward artificial neural network with a single hidden layer can approximate any continuous well-behaved function of arbitrary number of variables with any accuracy. The conditions are: a sufficient number of neurons in the hidden layer, and a correct weight selection. Above mentioned theorem for an arbitrary width case was originally proved by Cybenko \cite{cybenko1989approximation} and Hornik \cite{hornik1991approximation} and later extended to an arbitrary depth case (DNN) in \cite{lu2017expressive}. 

In this paper we get a deeper insight on the practical application of Cybenko’s theorem, in order to train a neural network, where all hidden neurons will be used efficiently. Therefore, we have to pay attention to two following aspects: number of neurons and correct weight selection.

\textit{Number of neurons} in a hidden layer is a quite straightforward parameter that became trendy with availability of multi-threaded parallel computing on GPU \cite{MARZIALE200773}. Models of a vast number of trainable parameters are not devoid of logic, as they generalize better and can be so-called ‘universal learners’. For example, GPT-3 having 175 billion parameters, is a perfect example of a universal learner \cite{brown2020language}. Thus, the community has been experimenting with model architectures increasing width \cite{lu2017expressive} or depth \cite{szegedy2014going} of neural networks. Issues, such as vanishing gradient \cite{Hochreiter:01book, hochreiter1998vanishing} was resolved by applying methods, including second-order Hessian-free optimization \cite{martens2011learning}, training schedules by using greedy layer-wise training \cite{schmidhuber1992learning, hinton2006fast, vincent2008extracting}, sparse rectifier activation function, widely known as ReLU \cite{glorot2011deep}, layer-size-dependent initialization, such as Xavier \cite{glorot2010understanding} and Kaiming \cite{he2015delving} and skip connections \cite{he2015deep}. Even though, we can make arbitrarily large models make good predictions, to achieve computational sustainability by expanding the number of trainable parameters up to infinity, would not be the best option for the tasks of lower complexity. The community has been already trying to address this problem, thus several solutions dealing with this issue have occurred. For example, widely used ReLU activation function, saturated only in one dimension, which helps with vanishing gradient problem, on the other hand results in so-called ‘dying neurons’ \cite{Lu_2020}, modified activation functions such as Leaky ReLU \cite{xu2015empirical}, adaptive convolutional ReLU \cite{gao2020adaptive}, Swish \cite{ramachandran2017searching}, Antirectifier \cite{luijten2019deep} and many other were addressed to solve the problem of ‘neural graveyard’. Resource efficient solutions, such as pooling operations \cite{scherer2010evaluation}, LightLayers \cite{jha2021lightlayers} depth-wise separable convolutions \cite{chollet2017xception} were developed to reduce the complexity of the models.

\textit{Correct weight selection}, at first sight, depends on training parameters, such as loss function, number of epochs, learning rate etc. However, to train the neural network competently these weights have to be initialized stochastically. There are several ways to initialize weight, mainly aimed to avoid vanishing gradients. Nevertheless, stochastic weight initialization can result in neuron redundancy, when different neurons are trained in a similar manner. This is not crucial if the neural network is excessively large, however, in computationally sustainable models, neuron redundancy and ‘neural graveyards’ are undesirable. Moreover, there are numerous application when memory efficient model is required (e.g. autonomous devices such as sensors, detectors, mobile or portable devices). Such devices require memory and performance efficient solutions to learn spontaneously and improve from experience. In this case adding excessive parameters to the model can be rather questionable for the model application.

Therefore, once we consider each neuron of the model as an individual learner, the neural network can be seen as an ensemble. It is known that for ensembles diversity of learners is desirable to some extent \cite{Brown04diversityin}. Thus, we can assume that diversity between neurons or reinforced diversification during the training can be beneficial for the model.

In this paper we foremost explore how the diversity between neurons evolves during the training and as a following step suggest methods for diversification of the neurons during the model training. This is especially relevant in resource constrained models, where neuron redundancy means reducing the number of predictors. Additionally, we show how weight pre-initialization can affect neural network training at the early steps.

\section{Our Approach}

Let us start with a term \textit{negative correlation} (NC) learning \cite{Brown04diversityin}, which is a simple, yet elegant technique to diversify individual base-models in the ensemble and reduce their correlations. Ambiguity decomposition \cite{hansen1990neural} of the loss function raises the possibility of controlling the trade-off between bias, variance, and covariance  \cite{ueda1996generalization} using the strength parameter, to reduce covariance. In its order the concept of an NC learning is originated from bias-variance decomposition \cite{bian2021does, izmailov2019averaging} of ensemble learning. In this case, bias is the output shift from the true value, and variance is the measure of ensemble ambiguity, which simply means dispersion around the mean output value.

As it was first demonstrated by Krogh and Vedelsby \cite{krogh1995validation} quadratic error of ensemble prediction is always less that the quadratic error of each individual estimator of the ensemble:

\begin{align}
\left(f_{\text {ens }}-d\right)^{2}=\sum_{i} w_{i}\left(f_{i}-d\right)^{2}-\sum_{i} w_{i}\left(f_{i}-f_{\text {ens }}\right)^{2}
\end{align}

Later Brown \cite{Brown04diversityin} demonstrated decomposition of ensemble error into three components - bias, variance and covariance, and shown, the connection between ambiguity and covariance:

\begin{equation}
\begin{array}{c}
E\left\{\frac{1}{M} \sum_{i}\left(f_{i}-d\right)^{2}-\frac{1}{M} \sum_{i}\left(f_{i}-\bar{f}\right)^{2}\right\}= \\
\\
 \overline{b i a s}^{2}+\frac{1}{M} \overline{v a r}+\left(1-\frac{1}{M}\right) \overline{\operatorname{covar}}
 \end{array}
\end{equation}

The ensemble ambiguity is nothing less than the variance of the weighted ensemble around the weighted mean. Therefore, higher ambiguity, i.e. decorrelation between the ensemble output is desirable up to some measure.

Our \textit{first trial} was to decorrelate neurons in the hidden layer by penalizing the difference between mean weight of the neurons $\bar{w}$ and each neuron $w_{i}$:

\begin{align}
NC = \frac{1}{\textit{n}}\gamma\sum_{i}^{}({\bar{w}}-{w_{i}})
\end{align} 
where $\gamma$ is the regularization strength parameter, and $\textit{n}$ is the number of neurons in a layer. 

However, it is likely more profitable to compare not only single weights, but weight matrices or e.g. kernels in convolutional neural networks (CNN), as trainable kernels represent. Thus, the \textit{second} way to define diversity is comparing neurons by cosine similarity:

\begin{align}
\frac{1}{{D}}=\frac{1}{\textup{n}}\gamma\sum_{i}^{}\sum_{j}^{} {w_{i}}\cdot{w_{j}}\
\end{align} 
where $\mathbf {w}$ are weights of individual neurons and $D$ is the diversity measure.

In this technique we compare each weight in the layers and define a diversity measure \(D\). However, it has quadratic complexity of such expression, which would oppose the idea of the current work, as our indent is fast and efficient training of resource constrained neural networks.

Therefore, combining the first two approaches we introduce and explore \textit{another method} to define diversity in the neural networks:

\begin{align}
\frac{1}{{D}}=\frac{1}{\textup{n}}\gamma\sum_{i}^{} {\bar{w}}\cdot{w_{i}}\
\end{align} 

After observing the training process and evolution of diversity measure in the models, we explored the possibility of weight pre-optimization using diversification. In this case, we used Kaiming weight initialization, with further optimization to enlarge the diversity between the weights, and at the same time keep weight mean and standard deviation of the weight matrix close to initial:

\begin{align}
L = (\left | \bar{\textup{W}} - \bar{w}_{k} \right |+ |\sigma_{\textup{W}} - \sigma_{w_{i}} |) \sum_{i}^{}\sum_{j}^{}{w_{i}}\cdot {w_{j}}\
\end{align} 
where $L$ is loss, $\bar{\textup{W}}$ is the initial weight mean, $\bar{w}_{k}$ is the weight mean at \textit{k} training step, $\sigma_{\textup{W}}$ is standard deviation of the initial weights array, and $\sigma_{w_{i}}$ is standard deviation of the weights array at \textit{k} training step.

\section{Experiments}
We perform some initial experiments using DNN in order to study diversity evolution during the model training and
demonstrate the effectiveness of proposed diversification mechanisms.

The experiments were performed on publicly available benchmark dataset Fashion MNIST \cite{xiao2017fashion}. This dataset was chosen as it is suitable for DNN training and has higher variance than traditional hand-written digits dataset MNIST \cite{lecun-mnisthandwrittendigit-2010}. We implemented one-hidden-layer neural network with 16, 32, 64, 128, and 256 neurons in the hidden layer (see Table 1), using PyTorch \cite{NEURIPS2019_9015} library. Otherwise, we used standard parameters for the training, including Adam optimizer \cite{kingma2017adam} with a learning rate of 0.01, cross entropy loss function with penalization terms (Eq. 3-5):

\begin{align}
H(T, p)=-\sum_{i=1}^{N} \frac{1}{N} \log _{2} q\left(x_{i}\right) + \frac{1}{{D}}
\end{align}
where $T$ presents training set, $p$ is true distribution, $q$ is predicted distribution, $N$ is standard deviation of the weights array at \textit{k} training step, $q(x)$ is the probability of event $x$ estimated from the training set, and ${D}$ is the diversity measure, obtained using Eq. 3, 4, or 5.

\section{Results and Discussion}

\subsection{Evolving Diversity and Symmetry Breaking}

During the model training, one can notice sub-optimal accuracy stagnation for a several epochs, this can be associated with the existence of local minima on a loss function surface \cite{10.1007/978-3-540-76928-6_12, swirszcz2017local}. This can be associated with a symmetry in the neural network layer, which is shown to be a critical point especially for small neural networks\cite{arjevani2020symmetry, tayal2020inverse}. We found out that naturally the model tends to decrease the correlation between the neurons, however when the model converges to a local minimum with a sub-optimal accuracy, the similarity between the neurons rises up until the moment when the optimization process surpasses the local minimum and the accuracy increases. (see Figure 1) This correlates with an existence of symmetry in the weights, once weights are symmetrical (correlated) and the number of neurons is constrained, the overall output of the model will likely to be inefficient.

\begin{figure}
  \includegraphics[width=.9\linewidth]{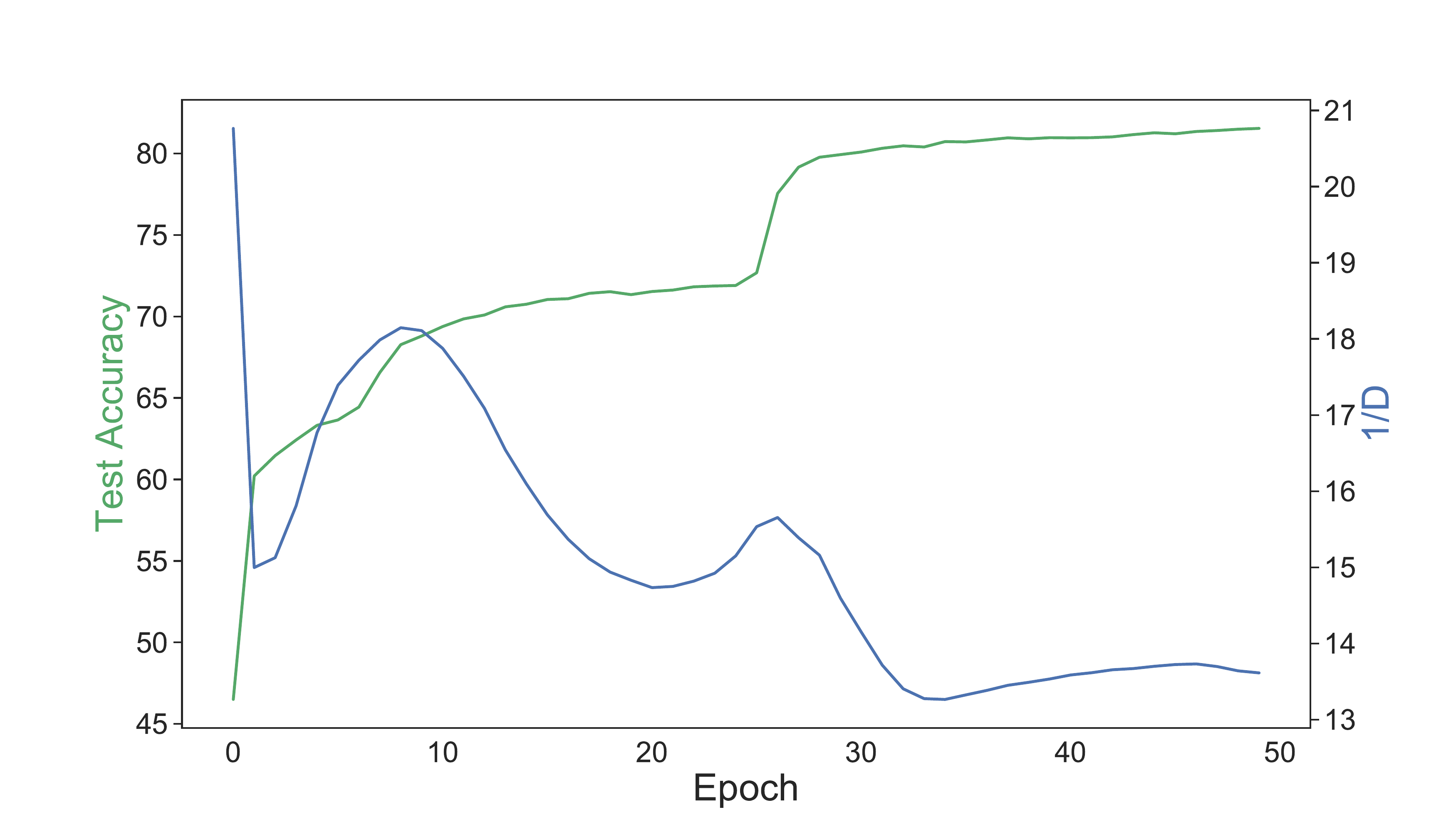}
  \caption{Training curve and Diversity measure (Eq. 4) for the first 50 epochs on Fashion MNIST dataset. DNN with 1 hidden layer of 32 neurons.}
  \label{fig:boat1}
\end{figure}

\subsection{Negative Correlation Learning}

\begin{table}[h!]
\centering
\caption{First 10 epochs average of the neural network training for various number of neurons, hidden layer diversified according to the Eq. 5.}
  \begin{tabular}{|c c c c c c|}
  \hline
    \multirow{2}{*}{$\gamma$} &
      \multicolumn{5}{|c|}{Number of Neurons \&} \\
    \multirow{2}{*}{} &
      \multicolumn{5}{|c|}{Test Accuracy, \%} \\
      \hline
      & {16} & {32} & {64} & {128} & {256} \\
      \hline
      \centering
    0.0 & 54.19 & 58.25 & 62.46 & 69.62 & 72.10\\
    \hline
    $1\cdot10^{-5}$ & 55.17 & 60.17& 62.45 & 68.64 & 70.65\\
    \hline
    $1\cdot10^{-4}$ & \textbf{56.41} & \textbf{61.25} & 64.13 & 70.32 &  72.27\\
    \hline
    $1\cdot10^{-3}$ & 54.48 & 60.81 & \textbf{65.04} & \textbf{70.45} & \textbf{72.83}\\
    \hline
    $1\cdot10^{-2}$ & 53.54 & 60.04 & 63.19 & 70.26 & 72.36\\
    \hline
    $1\cdot10^{-1}$ & 54.22 & 59.46 & 62.23 & 70.20 & 71.64\\
    \hline
    1.0 & 50.09 & 57.49 & 60.53 & 69.84 & 71.65\\
    \hline
  \end{tabular}
\end{table}

The experiment above inspired us to study certain ways to decorrelate neurons in the hidden layer, thus brake the symmetry that can appear during the learning process. As we discussed earlier, we consider the output of neural network as an output of an ensemble. Thus, first, we did simple NC learning, applied to the individual neurons, rather than ensemble of classifiers. The logic behind this experiment was rather comprehensible. Once the model has constrained number of parameters to generalize the data, higher variance would help to eliminate redundant neurons and overall prediction has to be more accurate. As it can be seen from the Figure 2. decorrelation mechanism helps to avoid local minima at the early stage on the model learning. Nevertheless, decorrelation using NC learning generally did not result in the higher accuracy overall. We associate it to several factors, such as Kaiming weight initialization that help to avoid vanishing gradient, and Adam optimizer which is a replacement optimization algorithm that can handle sparse gradients on noisy data, and thus is able to efficiently overcome local minima due to adaptive learning rated for each parameter. Eventhough, these widely used techniques are dealing with the above mentioned problem of the neuron redundancy, our proposed model can help at the early stages of a model training.

\begin{figure*}[t]
  \centering
  \subcaptionbox*{}[.3\linewidth][c]{%
    \includegraphics[width=.99\linewidth]{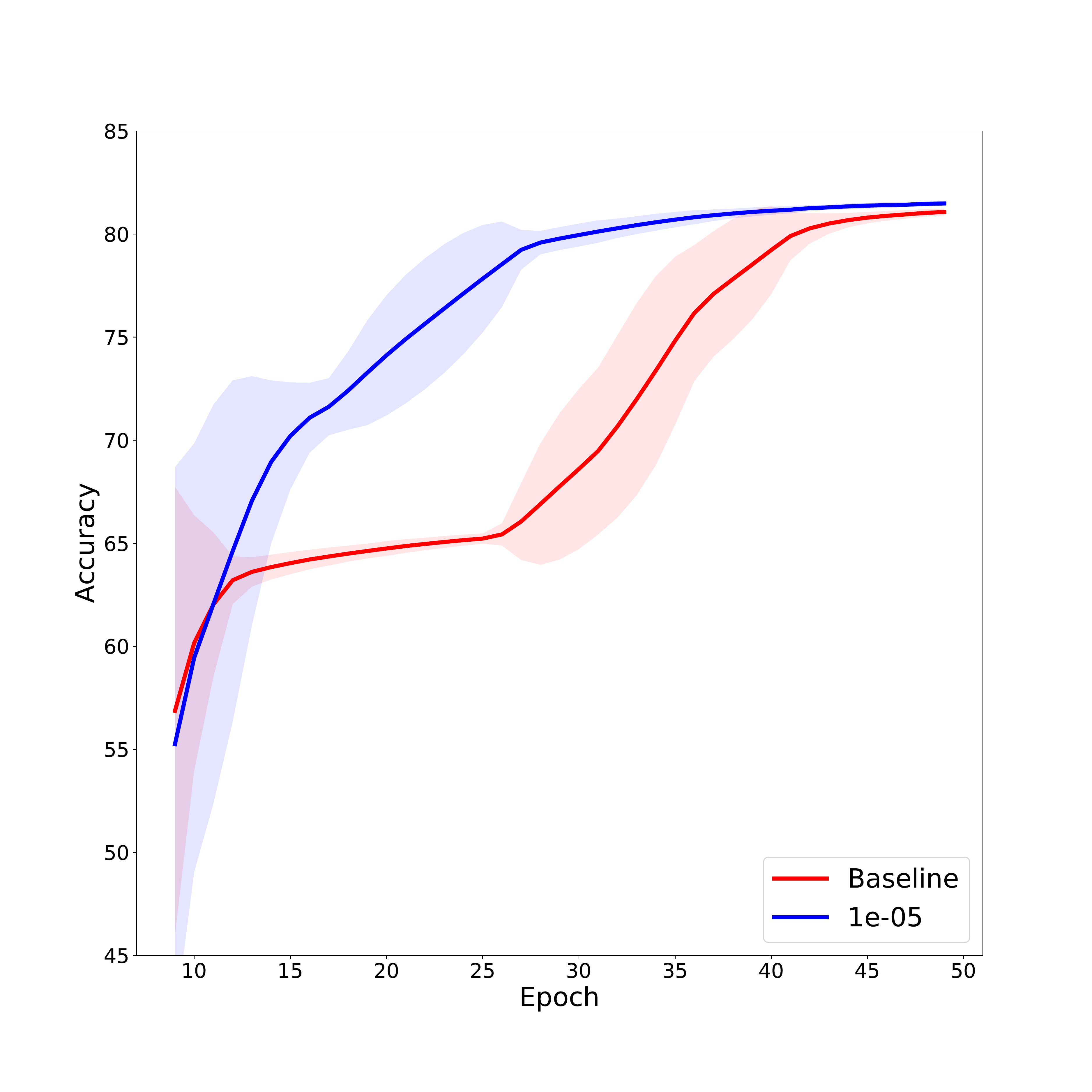}}
  \subcaptionbox*{}[.3\linewidth][c]{%
    \includegraphics[width=.99\linewidth]{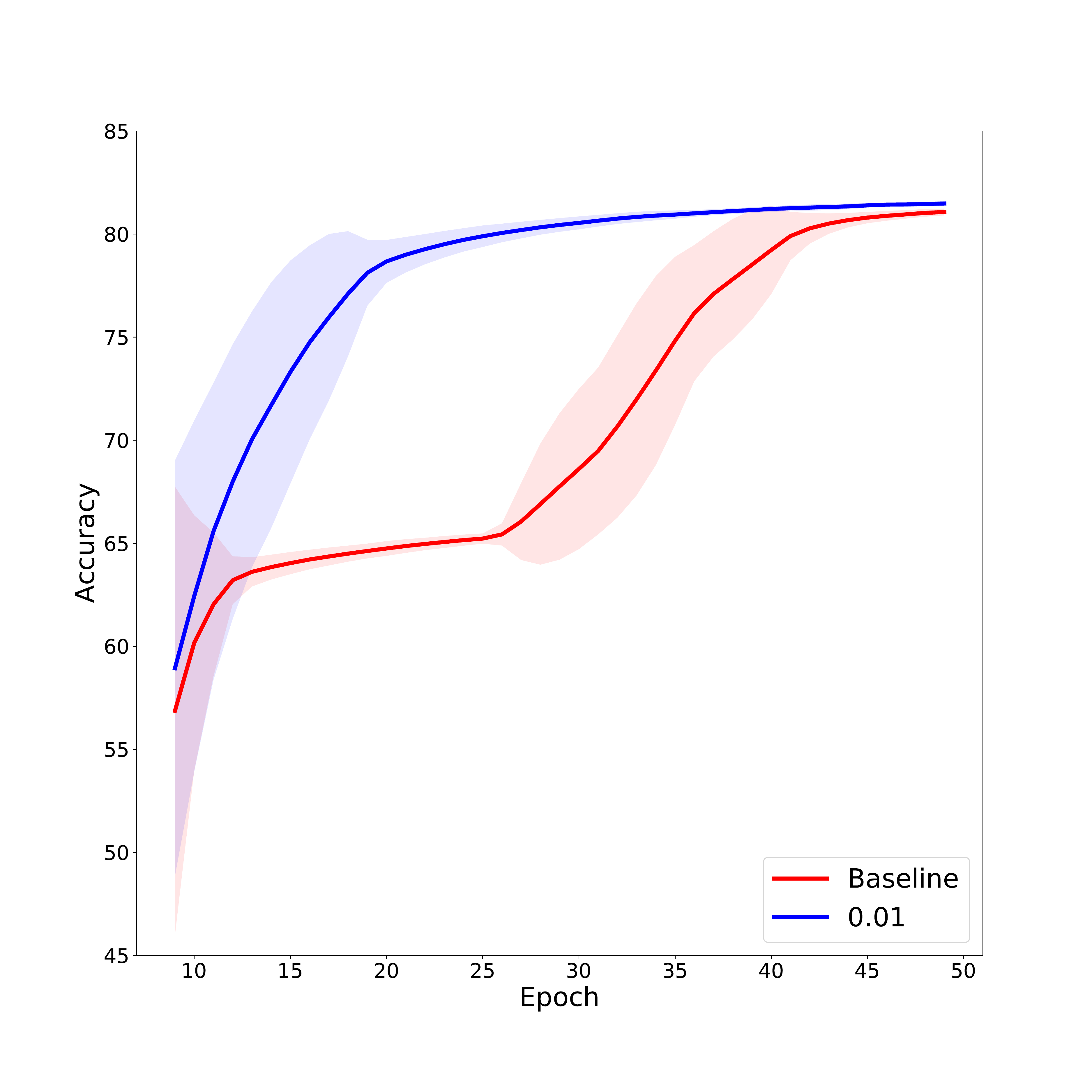}}
  \subcaptionbox*{}[.3\linewidth][c]{%
    \includegraphics[width=.99\linewidth]{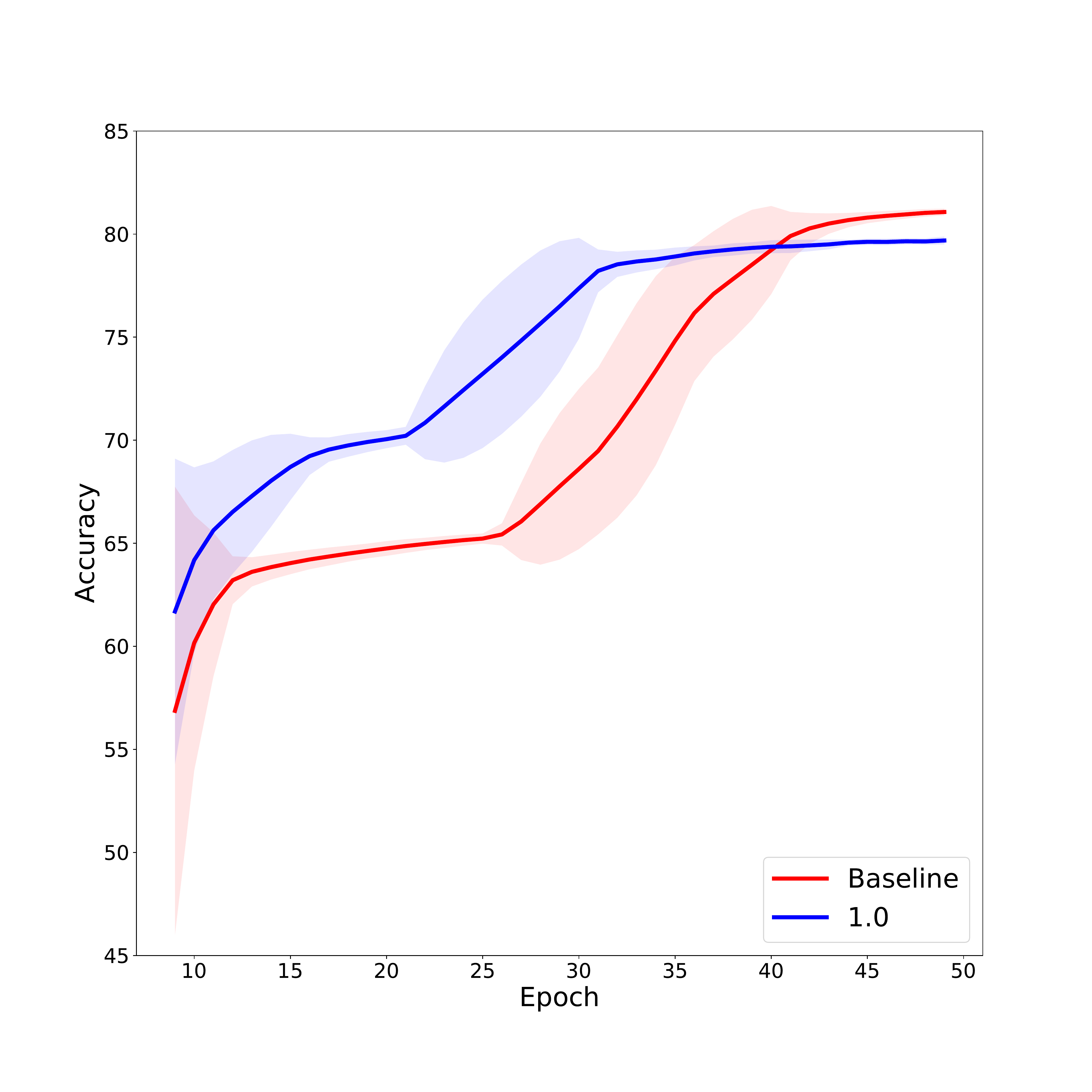}}

  \caption{Validation accuracy training curves of the model with various various $\gamma$ values.}
\end{figure*}

Moreover, with an increasing number of neurons the influence of decorrelation diminishes, this can be explained, that excessively large NN performs good at the low variance data as well as not every neuron is needed for a good prediction. However, in the present work we consider computationally sustainable DNN, where all the neuron are forced to contribute the prediction and on the other hand, for complex data larger amount of neurons would be needed to generalize the dataset. Therefore, for more sophisticated problems neuron diversification may be efficient for a larger number of neurons. However, in the present case we performed further experiments on the model with 64 neurons in the hidden layer, which we consider sufficient for a given dataset. All the models were trained for 10 times to calculate mean and standard deviation. In Table 2 the average testing accuracy of the first 10 epoch for the DNN with 64 neurons in the hidden layer trained using negative correlation learning (Eq. 3) is shown.

\begin{table}[h!]
\begin{center}
\caption{First 10 epochs average of the neural network training, hidden layer diversified according to the Eq. 3.}
 \begin{tabular}{|c c c c|}
 \hline
 {$\gamma$} & {Train Acc., \%}  & {Test Acc., \%} , & {Test Acc. STD} \\ 
 \hline
0.0 & 61.46 & 62.46 & 2.34 \\ 
 \hline
$1\cdot10^{-5}$ & 62.26 & 62.45 & 2.34 \\
 \hline
$1\cdot10^{-4}$ & 63.65 & 64.13 & 1.59 \\
 \hline
$1\cdot10^{-3}$ & 63.12 & \textbf{65.04} & 1.76 \\
 \hline
$1\cdot10^{-2}$ & 62.54 & 63.19 & 1.03\\ 
 \hline
$1\cdot10^{-1}$ & \textbf{64.23} & 62.23 & 0.95 \\ 
 \hline
1.0 & 59.56 & 60.53 & 1.57 \\
 \hline
\end{tabular}
\end{center}
\end{table}

\subsection{Pairwise Cosine Similarity Diversification}

It has to be noted that, unlike in \cite{Brown04diversityin}, where universal diversification strength parameter was found for the ensembles of all sizes, in our case $\gamma$ value depends on the size of the hidden layer and has to be rather considered as \textit{$\gamma$ per neuron}. However, on the other hand it is loss-dependent, which means that, ideally, it has to be same or one order of magnitude smaller than the output of the loss function during the training, otherwise, rather than the model loss (e.g. cross entropy), reciprocal diversity measure \(\frac{1}{D}\) will be optimized. Thus, the reader has to consider optimizing $\gamma$ value for each certain neural network and loss function. Thus optimal $\gamma$ approximately can be estimated as:

\begin{align}
\gamma = \frac{0.5 \cdot 10^{b_{loss}}}{\textup{n}}
\end{align} 
where $\textup{n}$ is the number of neurons in the hidden layer and ${b_{loss}}$ is the loss function order of magnitude.

In addition to NC learning, we introduced diversity measure based on cosine similarity between the neurons (Eq. 4). Such technique, seems to be promising due to several reasons: first, we, rather that mean values, compare patterns, which can be useful for more complex models, such as CNNs or transformers, moreover here, each neuron is compared with each, thus such model is intended to be more robust. Nevertheless, at least for DNN, results we comparable with NC learning (see Table 3), additionally, such method has quadratic complexity, which opposes our initial aim to train small models faster and more efficient.

\begin{table}[h!]
\begin{center}
\caption{First 10 epochs average of the neural network training, hidden layer diversified according to the Eq. 4.}
 \begin{tabular}{|c c c c|}
 \hline
 {$\gamma$} & {Train Acc., \%}  & {Test Acc., \%} , & {Test Acc. STD} \\
 \hline\hline
0.0 & 55.94 & 62.83 & 1.18 \\ 
 \hline
$5\cdot10^{-8}$ & 56.52 & 64.20 & 1.11 \\
 \hline
$5\cdot10^{-7}$ & \textbf{57.98} & \textbf{65.76} & 0.92 \\
 \hline
$5\cdot10^{-6}$ & 55.48 & 59.96 & 0.71 \\
 \hline
$5\cdot10^{-5}$ & 56.47 & 49.20 & 1.52 \\ 
 \hline
$5\cdot10^{-4}$ & 56.47 & 44.60 & 1.06 \\ 
 \hline
$5\cdot10^{-3}$ & 42.66 & 38.61 & 1.10 \\ 
 \hline
\end{tabular}
\end{center}
\end{table}

\subsection{Reaching Linear Complexity}

To enable our diversification method to compare patterns, however avoid quadratic complexity, we combined the fist concept of NC learning with the second one, and implemented diversity measure based on penalization of the cosine similarity of each neuron in the hidden and layer's neurons mean (Eq. 5). The algorithm (see Table 4) overhead is comparable with $L$ regularization. Moreover, it has shown the highest accuracy gain among three.

\begin{table}[h!]
\begin{center}
\caption{First 10 epochs average of the neural network training, hidden layer diversified according to the Eq. 5.}
 \begin{tabular}{|c c c c|}
 \hline
{$\gamma$} & {Train Acc., \%}  & {Test Acc., \%} , & {Test Acc. STD} \\
 \hline
0.0 & 61.54 & 63.15 & 2.08 \\ 
 \hline
$5\cdot10^{-7}$ & 62.37 & 63.3 & 1.63 \\
 \hline
$5\cdot10^{-6}$ & 63.60 & 64.54 & 1.25 \\
 \hline
$5\cdot10^{-5}$ & \textbf{64.87} & \textbf{64.95} & 1.66\\
 \hline
$5\cdot10^{-4}$ & 60.36 & 62.14 & 1.66 \\ 
 \hline
$5\cdot10^{-3}$ & 52.54 & 55.86 & 0.45 \\ 
 \hline
$5\cdot10^{-2}$ & 41.26 & 42.71 & 1.32 \\ 
 \hline
\end{tabular}
\end{center}
\end{table}

\subsection{Iterative Diversified Weight Initialization}

However, it can be noticed, that occasionally, during the training, the model do not behave exactly as expected, creating an outlying learning curves. This is most likely associated with stochastic weight initialization. In this case Kaiming initalization is used \cite{he2015deep}. Kaiming initialization is widely used for the neural networks with ReLU activation functions and related to the nonlinearities of the ReLU activation function, which make it non-differentiable at \(x=0\). The weights, in this case are initialized stochastically with the variance that depends on the number of neurons \(N\):

\begin{align}
v^{2} = 2/N
\end{align} 

It is fair to suggest, that correlation between the initialized weights can play significant role in the model learning process. Indeed, in the Figure 1. it is clearly seen, the the model gained the most of its accuracy while reducing the correlation between neurons during the first few epochs. However, the aim of weight initialization is to prevent layer activation outputs from exploding or vanishing during the course of a forward pass through a deep neural network. Usually weight are initialized stochastically with a small number to avoid vanishing gradients especially if \(tanh\) or \(sigmoid\) activation functions are used. Thus, to obtain stochastically initialized, yet decorrelated, weights we introduced iteratively diversified Weight initialization, using custom loss function based on Eq. 6. The logic behind such initialization is to reduce the diversity measure between the weights and at the same time keep weights mean \(\bar{w}\) and weights standard deviation \(\sigma_{w}\) close to the originally initialized using Kaiming initialization.

\begin{table}[h!]
\begin{center}
\caption{First 5 epochs average of the neural network training initialized with decorrelated weights according to the Eq. 6 pre-optimized for 5 epochs.}
 \begin{tabular}{|c c c c|}
 \hline
  {$\gamma$} & {Train Acc., \%}  & {Test Acc., \%} , & {Test Acc. STD} \\
 \hline
0.0 & 29.54 & 34.23 & 2.04 \\ 
 \hline
$1\cdot10^{-4}$ & 42.43 & 43.43 & 1.81 \\
 \hline
$1\cdot10^{-3}$ & 43.92 & 45.65 & 1.53 \\
 \hline
$1\cdot10^{-2}$ & \textbf{44.65} & \textbf{47.01} & 1.24 \\
 \hline
$1\cdot10^{-1}$ & 38.32 & 39.83 & 1.06 \\ 
 \hline
1.0 & 36.64 & 38.94 & 1.37 \\ 
 \hline
10.0 & 32.5 & 37.57 & 1.41 \\ 
 \hline
\end{tabular}
\end{center}
\end{table}

\section{Conclusion}
In this paper we show how to explore and tame the diversity of neurons in the hidden layer. We studied how the correlation between the neurons evolves during the training and what is the effect on prediction accuracy. In appears, that once the model is converged to the local minimum on the loss landscape, correlation between the neurons increases up to the point when the optimization process overcome the local minimum. Thus, we introduced three methods how to dynamically reinforce diversification and thus decorrelate neural network layer. The concept of negative correlation suggested by Brown \cite{Brown04diversityin} was reviewed and expanded. Instead of decorrelation individual neural networks in the ensemble we diversified neurons in the hidden layer, using three techniques: \textit{negative correlation learning, cosine pairwise similarity, cosine similarity around the mean}.

First technique is originated from the neural networks ensembles and shows a decent performance in our example using DNN, however for more sophisticated models, such as CNNs and transformers, second and third technique is likely to be more advantageous as far as it can compare patterns. Additionally to reach correct weight selection, we introduced weight iterative optimization using weight diversification. It was shown that such techniques are suitable for the fast training of small models and notably affect their accuracy at the early stage. Which is a small, yet important step towards the development of a strategy towards energy-efficient training of neural networks.

Our future plans for using neural network diversification primarily consists in using above described diversification techniques in more sophisticated models in order to explore the possibility to improve training speed and reduce the number of training parameters. Popular architectures, such as transformers can benefit from the individual head diversification in multi-head attention block, as far as multiple heads are intended to learn various representation. Furthermore, we are planning to explore more pattern-oriented techniques for defining diversity between neurons to enable efficient diversification application in CNNs.

\section*{Acknowledgment}
This research is supported by the Czech Ministry of Education, Youth and Sports from the Czech Operational ProgrammeResearch, Development, and Education, under grant agreement No. CZ.02.1.01/0.0/0.0/15003/0000421 and the Czech Science Foundation (GA\v{C}R 18-18080S).
%
%
%
\bibliography{arxiv}{} 
\bibliographystyle{unsrt}
\end{document}